%% file: conference_101719.tex
\documentclass[conference]{IEEEtran}
\IEEEoverridecommandlockouts
\usepackage{cite}
\usepackage{amsmath,amssymb,amsfonts}
\usepackage{algorithmic}
\usepackage{graphicx}
\usepackage{textcomp}
\usepackage{xcolor}

\usepackage{url}
\usepackage{array}
\usepackage{booktabs}
\usepackage{colortbl} 
\usepackage{algorithmic}
\usepackage{textcomp}

\usepackage{booktabs}
\usepackage{comment}
\usepackage{hyperref} 
\usepackage{caption}
\usepackage{makecell}

\newenvironment{dany}{\color{red}}{\color{black}}
\newenvironment{hamza}{\color{blue}}{\color{black}}

\def\BibTeX{{\rm B\kern-.05em{\sc i\kern-.025em b}\kern-.08em
    T\kern-.1667em\lower.7ex\hbox{E}\kern-.125emX}}
\begin{document}

\title{Unsupervised Deep Learning-based clustering for Human Activity Recognition}

\author{\IEEEauthorblockN{Hamza Amrani, Daniela Micucci and Paolo Napoletano}
\IEEEauthorblockA{\textit{Department of Informatics, Systems and Communication} \\
\textit{University of Milano - Bicocca}
, Milan, Italy \\
Email: h.amrani@campus.unimib.it, daniela.micucci@unimib.it, paolo.napoletano@unimib.it}
}
\maketitle

\begin{abstract}
One of the main problems in applying deep learning techniques to recognize activities of daily living (ADLs) based on inertial sensors is the lack of appropriately large labeled datasets to train deep learning-based models. 
A large amount of data would be available due to the wide spread of mobile devices equipped with inertial sensors that can collect data to recognize human activities. Unfortunately, this data is not labeled.
The paper proposes DISC (Deep Inertial Sensory Clustering), a DL-based clustering architecture that automatically labels multi-dimensional inertial signals. In particular, the architecture combines a recurrent AutoEncoder and a clustering criterion to predict unlabelled human activities-related signals. 
The proposed architecture is evaluated on three publicly available HAR datasets and compared with four well-known end-to-end deep clustering approaches. The experiments demonstrate the effectiveness of DISC on both clustering accuracy and normalized mutual information metrics.

\end{abstract}
\begin{IEEEkeywords}
Human Activity Recognition, Unsupervised learning, Deep learning, AutoEncoder
\end{IEEEkeywords}


\input{sections/introduction}
\input{sections/architecture}
\input{sections/material_and_methods}
\input{sections/results}
\input{sections/conclusions}


\bibliographystyle{IEEEtran}
\bibliography{bibliography}

\end{document}

%% file: sections/introduction.tex
\section{Introduction and Background}
\label{section:introduction}

Automated recognition of activities of daily living (ADLs) based on inertial data is popular in various fields, such as surveillance, health care, and delivery~\cite{sun2020using,mukherjee2020ensemconvnet,iyengar2020covid}. 
Nowadays, supervised classification of ADLs is mainly based on deep learning (DL) approaches, which are well known to be data hungry~\cite{ferrari2021trends}. Many labeled datasets are available in the literature, however to scale-up the use of these methods in several domains, a very large amount of data is required~\cite{amrani2022homogenization}. 

One way to speed up the labeling of ADL samples is to take advantage of the clustering analysis that is an unsupervised strategy to group similar data into clusters~\cite{abedin2020towards}. Its applications include automatic data labeling for supervised learning and pre-processing for data visualization and analysis.

Most common methods consist of a two-stages process: \emph{stage 1)} a deep neural network (DNN) procedure for learning deep representative features and \emph{stage 2)} a Machine-Learning (ML)-based clustering algorithm for data grouping~\cite{min2018survey,gan2020data,10.1093/bib/bbz170}.
Deep Embedded Clustering (DEC)~\cite{xie2016unsupervised} learns a mapping from the observed space to a low-dimensional latent space leveraging Stacked AutoEncoders (SAE), which can obtain feature representations and cluster assignments simultaneously.

%

Improved Deep Embedded Clustering (IDEC)~\cite{guo2017improved} is a modified version of DEC that incorporates an under-complete AutoEncoder to preserve the local structure. The preservation of the local structure prevents the embedded space from being distorted by fine-tuning, thus ensuring embedded spatial features’ representativeness.


Deep Convolutional Embedded Clustering (DCEC)~\cite{guo2017deep} improves IDEC by replacing Stacked AutoEncoders (SAE) with Convolutional AutoEncoders(CAE) to preserve the local data structure. 


Recently, Balanced Deep Embedded Clustering (BDEC)~\cite{obeid2021unsupervised} has been proposed to address the inherent vulnerability of DEC to data imbalance by utilizing a pre-processing step that uses a scalable representative algorithm to extract a balanced subset and use it for training.

Finally, Deep Sensory clustering (DSC)~\cite{abedin2020towards} is a recent clustering architecture that employs a Recurrent AutoEncoder (RAE) and a clustering criterion to learn clustering-friendly representations from inertial sensory readings. DSC can be considered the first approach using deep clustering on inertial signals.



The paper proposes Deep Inertial Sensory Clustering (DISC), a deep-learning-based clustering architecture that automatically performs unsupervised learning and label annotation by analyzing multi-dimensional inertial signals. 
The architecture combines a recurrent AutoEncoder and a clustering criterion to predict unlabelled human activities-related signals.
As for previous approaches in the state of the art, our architecture consists of two parts. The first part covers stage 1. It consists of an AutoEncoder that tries to reconstruct two outputs from the input signals: the inverse input and the future sequence of the input. The second part covers stage 2 and is devoted to clustering the dimensionally reduced signals from stage 1.

The main contribution of the paper with respect to the state of the art is the use of ConvGRU layers (Convolutional Gate Recurrent Unit~\cite{ballas2015delving}) that allow spatio-temporal features to be extracted. The recursive layers (LSTM, GRU) have been shown to be useful in extracting temporal features, 
while the convolutional layers in extracting spatial features. 

The achieved architecture has been compared with four state-of-the-art techniques (DEC~\cite{xie2016unsupervised}, IDEC~\cite{guo2017improved}, DCEC~\cite{guo2017deep}, and DCS~\cite{abedin2020towards}) on three publicly available HAR datasets (UCI HAR~\cite{anguita2013public}, Skoda~\cite{stiefmeier2008wearable}, and MHEALTH~\cite{banos2014mhealthdroid}). Results show, on average,  its effectiveness in the inertial sensory clustering task.

The paper is organized as follows. Section~\ref{section:architecture} introduces the architecture; Section~\ref{section:material_and_methods} describes the implementation details of the architecture, the metrics used to validate the approach, and the datasets used in the evaluation. Section\ref{section:results} discusses the results obtained. Finally Section~\ref{section:conclusions} presents the conclusions and the future directions.
    
\begin{hamza}    

\end{hamza}

%% file: sections/architecture.tex
\section{Architecture}
\label{section:architecture}

\begin{figure*}[h]
  \centering
  \includegraphics[width=1.0\textwidth]{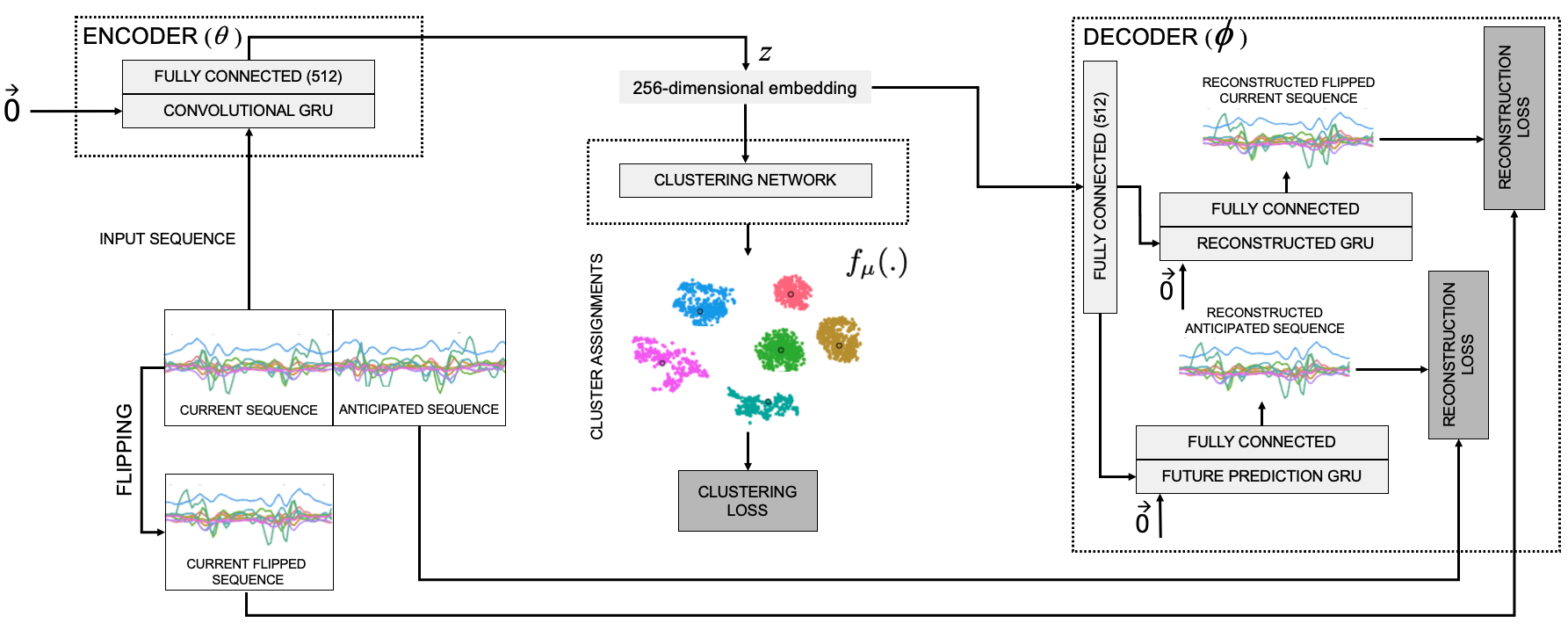}
  \caption{Deep Learning-based Inertial Sensor Clustering architecture.}
  \label{fig:architecture}
\end{figure*}

This section presents our proposal Deep Inertial Sensory Clustering (DISC). The neural architecture of our proposal is sketched in Figure~\ref{fig:architecture}. The first subsection shows the architecture used to implement stage 1 (i.e., the Multi-Task AutoEncoder employed), and the second the architecture for stage 2 (i.e., the Clustering Criterion adopted).

\subsection{Stage 1: Multi-Task AutoEncoder}

The architecture leverages a \emph{Recurrent Neural Network (RNN)-based autoencoder (RNN-AE)} since RNN architectures have been proved to be powerful for handling sequential data, such as text, sounds, and HAR signals~\cite{graves2013hybrid,murad2017deep}. 

The RNN-AE architecture consists of three RNNs: a recurrent encoder ConvGRU~\cite{ballas2015delving} ($Enc_{\theta}$) and two conditional decoders GRU~\cite{cho2014learning} ($Dec_{\phi}$). 

The input is a window $x$ of length $T$ which contains $x_{1}, ..., x_{T}$ elements. Each $x_{t}$ element includes one or more values according to the number of channels $N$.

\subsubsection{Recurrent Encoder ($Enc_{\theta}$)}

The recurrent encoder $Enc_{\theta}$ learns a compacted representation of the spatio-temporal features that characterizes the activities of daily living (ADLs).

In particular, a bi-directional Convolutional Gated Recurrent Unit (ConvGRU)~\cite{ballas2015delving}\footnote{A RNN that combines Gated Recurrent Units (GRUs)~\cite{cho2014learning} with the convolution operation.} reads a windows $x$ in both forward and backward directions and updates its hidden internal state for each $x_{t} \in x$. 

Equations from~\ref{eq:rule-begin} to~\ref{eq:rule-end} define the update rule.

\begin{equation} 
g_t = \sigma(W_g \star_m [h_{t-1};x_t]+b_g )
\label{eq:rule-begin}
\end{equation}
\begin{equation} 
r_t = \sigma(W_r \star_m [h_{t-1};x_t]+b_r )
\end{equation}
\begin{equation} 
\tilde{h_t} = tanh (W_c \star_m [x_t; g_t\odot h_{t-1} ] + b_c)
\end{equation}
\begin{equation} 
h_t = (1-r_t)\odot \tilde{h_t} + r_t \odot h_{t-1} 
\label{eq:rule-end}
\end{equation}

where $h_{t}$ is the output for $x_{t-1}$; $g_t$ and $r_t$ are the \emph{update} and the \emph{forget} gates respectively, $\sigma$ is the \emph{sigmoid} function, $\star_m$ represents a \emph{convolution} with a kernel of size $m\;$x$\;m$, $b$'s are \emph{bias terms}, $\odot$ denotes \emph{element-wise multiplication (or Hadamard product)}, and $W_{g,r,c}$ are 2D-convolutional kernels. 




After processing the entire sequence $x$, the final hidden state $h_{T}$ of 512 dimensions is reduced through a fully connected layer to 256 dimensions. 

The resulting low-dimensional feature $z \in \mathbb{R}^d$ of equation~\ref{eq:enc} encodes information about the ADL through a representation of the spatio-temporal dependencies that are present in the input sequence $x$. The size of the feature space is $d = 256$. 

\begin{equation} 
z = Enc_{\theta}(x)
\label{eq:enc}
\end{equation}

The hidden states of the convolutional GRU is initialized with zeros. 

\subsubsection{Conditional Recurrent Decoders ($Dec_{\phi}$)}

The decoder is based on gated GRU and it is in charge of reconstructing back the input sequence $x$ from $z$ by using a multi-task strategy that includes two different decoders $Dec_{\phi}$. This strategy minimizes two different Mean Square Error (MSE) losses: one to optimize the reconstruction of a time-flipped version of the input sequence, and the other to optimize the reconstruction of a future (anticipated) sequence. 

In the proposed architecture, both recurrent decoders are conditional. In fact, the decoder GRU needs two inputs at stage $t$: the input sequence $x_{t}$ and the hidden state $h_{t}$. At the decoder stage we do not have an input sequence and thus the input $x_{t}$ is initialized, for $t=0$, with zeros. Differently,  the hidden state is initialized at stage $t=0$ with a transformation of the embedding $z$. This transformation, named \emph{context vector}, is achieved through a convolutional layer that maps a 256-dimensional embedding in a 512-dimensional one. At stage $t>0$ the input $x_{t} = h_{t-1}$ while $h_{t}$ is the hidden state at previous stage that is updated by the training procedure.  

This multi-task strategy, along with the conditional initialization of the hidden states, forces the AutoEncoder to learn a meaningful representation (embedding space) where  sequences  belonging to different classes are well separated in the representation space.  


Equations from~\ref{eq: activation-begin} to~\ref{eq:activation-end} define the activation $h_t$.


\begin{equation} 
g_t = \sigma(W_g x_t + U_g h_{t-1})
\label{eq: activation-begin}
\end{equation}
\begin{equation} 
r_t = \sigma (W_r x_t + U_r h_{t-1}))
\end{equation}
\begin{equation} 
\tilde{h_t} = tanh (W x_t + U(r_t \odot h_{t-1}))
\end{equation}
\begin{equation} 
h_t = (1-g_t)h_{t-1} + g_t \tilde{h_t}
\label{eq:activation-end}
\end{equation}


Summing up, the output of the decoder is:
\begin{equation} 
(\bar{y}^{\;rec},\bar{y}^{\;fut})=Dec_{\phi}(z)
\end{equation}
where $\bar{y}^{\;rec}$ and $\bar{y}^{\;fut}$ are the reconstructed and the anticipated sequences generated from the input $x$.


The \emph{objective} of the Recurrent AutoEncoder is a joint objective function:
\begin{equation}\begin{split}
L_{AE} =  L_{rec} + L_{fut} 
= \\
    \left\| y^{\;rec} - \bar{y}^{\;rec} \right\|^2 + \left\| y^{\;fut} - \bar{y}^{\;fut} \right\|^2
\end{split}\end{equation}

where $L_{rec}$ and $L_{fut}$ indicate the reconstruction loss and the future prediction loss, respectively, and denote the mean square errors between decoder's generated output sequences ($\bar{y}^{\;rec}$ and $\bar{y}^{\;fut}$ ) and the expected target sequences ($y^{\;rec}$ and $y^{\;fut}$).

The \emph{optimal network parameters} of encoder $z=Enc_{\theta}(x)$ and decoder $(\bar{y}^{\;rec},\bar{y}^{\;fut})=Dec_{\phi}(z)$ are updated by minimizing the reconstruction error:
\begin{equation} 
(\theta^*, \phi^*) =  min_{\theta, \phi} \; L_{AE}
\end{equation}

\subsection{Stage 2: Clustering Criterion}
A parameterized clustering network $f_{\mu}(.)$ is connected to the AutoEncoder embedding layer to estimate the cluster assignment distributions and to map each $z$ into a soft label. 


It uses the similarities between the data representations and cluster centroids $\{\mu_j\}_{j=1}^k$  to compute soft cluster assignments, while its loss enforces the soft assignments to have more stringent probabilities.
The initial cluster centroids are obtained from classical Clustering algorithms on the embedded representations $z=Enc_{\theta}(x)$ after the pre-training of Multi-Task AutoEncoder, and are initialized only once at the beginning of the refinement stage. 

The clustering network $f_{\mu}(.)$ mantains cluster centroids $\{\mu_j\}_{j=1}^k$ as trainable weights and maps a set of $m$ embedded points $\{ z_i \in Z \}^m_{i=1}$ into soft label $Q_i = f_{\mu}(z_i) = (q_{ij})^k_{j=1}$ by following the Student's $t$-distribution~\cite{van2008visualizing}:

\begin{equation} 
q_{ij} = \frac{ (1 + \left\| z_i - \mu_j \right\|^2)^{-1} }{ \sum_{j'=1}^{k} (1 + \left\| z_i - \mu_{j'} \right\|^2)^{-1} }
\end{equation}
where  $q_{ij}$ is the $j$-th entry of $q_i$, which represents the probability of $z_i$ belonging to cluster $j$.

By squaring this distribution and then normalizing it, the auxiliary distribution $P_i = (p_{ij})^k_{j=1}$~\cite{xie2016unsupervised} forces assignments to have stricter probabilities (i.e., closer to 0 and 1). $P_i$ helps to improve cluster purity, emphasizing on data points assigned with high confidence, and to prevent large clusters from distorting the hidden feature space.\\
It is defined as:
\begin{equation}
p_{ij} = \frac{ q_{ij}^2 / \sum_i^m q_{ij}}{\sum_{j'=1}^{k} ( q_{ij'}^2 / \sum_{i=1}^m q_{ij'} )  }
\end{equation}

The \emph{Cluster Assignment Hardening (CAH)} loss $L_C$ is defined through minimizing the \emph{Kullback-Leibler (KL) divergence}~\cite{kullback1951information} between the distribution of soft labels and the auxiliary target distribution, as:
\begin{equation}
L_C = KL (P || Q) = \sum_i^m \sum_j^k p_{ij} \; log \frac{p_{ij}}{q_{ij}}
\end{equation}
Lower the KL divergence value, the better we have matched the true distribution with our approximation.

The reconstruction loss of the AutoEncoder is joined to the objective and optimized along with the Cluster Assignment Hardening loss simultaneously, preserving the local structure of data generating distribution and avoiding the corruption of feature space. The final joint optimization criterion is:
\begin{equation} 
 L =  \gamma L_{C} + L_{AE}
\end{equation}
where the coefficient $\gamma \in [0,1]$ controls the Clustering objective contribution.
The optimal network parameters are optimized with respect to the global criterion as:
\begin{equation} 
(\theta^*, \phi^*, \omega^*) =  min_{\;\theta, \phi, \mu} \;  L
\end{equation}

%% file: sections/material_and_methods.tex
\section{Material and Methods}
\label{section:material_and_methods}

\subsection{Network Architecture}

The recurrent encoder $Enc_{\theta}$ is a 2-layer bi-directional ConvGRU with 256 hidden states 
and the conditional recurrent decoders $Dec_{\phi}$ use uni-directional connections with 512 hidden states. The bottleneck embedded dimension of autoencoders is set to 256. 

A single layer in the clustering network $f_{\mu}(.)$ has been used to integrate the CAH (Clustering Assignment Hardening).

All experiments share the same network architecture.

\subsection{Optimization Settings}

The recurrent AutoEncoder is pre-trained end-to-end in mini-batches of size 256 for 100 epochs using the Adam optimizer~\cite{kingma2014adam}. The initial learning rate has been set to 10 and, after 70 epochs, decayed by a factor of 10. 

A Batch Normalization~\cite{ioffe2015batch} operation is used in the encoder fully connected layer to make the training faster and more stable.
Then, the recurrent AutoEncoder is refined with the clustering objective till the cluster assignment changes among two consecutive epochs are less than 0.1\%. 

The coefficient $\gamma$ is set to 0.1. The cluster centroids are initialized with $k$-Means and agglomerative clustering algorithms.

The parameters are maintained constant in all the experiments.



\subsection{Baseline}

Four state-of-the-art approaches served as baselines.

\subsubsection{Deep Embedded Clustering (DEC)~\cite{xie2016unsupervised}} relies on  AutoEncoders as initialization method and uses $k$-Means for clustering. First, the method pre-trains the model using an input reconstruction loss function (non-clustering loss). Second, a clustering algorithm initializes the clustering centers, without considering the decoder. Then, the model is optimized using the Cluster Assignment Hardening (CAH) loss, and the clustering centers are updated at each iteration by minimizing the Kullback-Leibler (KL) divergence as a loss function.

\subsubsection{Improved Deep Embedded Clustering (IDEC)~\cite{guo2017improved}} is a modified version of DEC that preserves the decoder layer, which is used for clustering loss. The reconstruction
loss of AutoEncoders is added to the objective and optimized simultaneously with clustering loss, preserving the local data structure.

\subsubsection{Deep Convolutional Embedded Clustering (DCEC)~\cite{guo2017deep}} uses the IDEC algorithm by incorporating a Convolutional AutoEncoder (CAE). Due to its spatial mapping relationships, CAE is able to learn deeper and spatial embedded features for clustering.

\subsubsection{Deep Sensory Clustering (DSC)~\cite{abedin2020towards}} is an alternative to IDEC. It is based on a recurrent AutoEncoder with two different decoders: one for input sequence reconstruction and one for anticipation sequence reconstruction. By using two decoders the AutoEncoder is forced to learn highly discriminative embedded features that are used to initialize the hidden state of the decoders.

\subsection{Datasets}
The proposed architecture and baselines were evaluated on three datasets.


\subsubsection{Human Activity Recognition Using Smartphones Dataset (UCI HAR)~\cite{anguita2013public}} includes 3-axial linear acceleration, 3-axial angular velocity, and gyroscope sensor data. The signals were recorded with a Samsung Galaxy S II smartphone at a constant rate of 50Hz, and the activities were performed by 30 volunteers. 
Each subject performed 6 activities. 
All the participants were wearing a smartphone on the waist during the experiment execution.
    
\subsubsection{Skoda Mini Checkpoint Activity Recognition Dataset (Skoda)~\cite{stiefmeier2008wearable}} includes 20 sensors (3-axial acceleration) placed on the left and right upper and lower arm. The signals were recorded at a sampling rate of 98Hz, and the activities were performed by 1 volunteer who executed 19 times each activity. The dataset contains 10 activities 
    
\subsubsection{mHealth Dataset (MHEALTH)~\cite{banos2014mhealthdroid}} includes 3-axial acceleration from the chest sensor, 2 electrocardiogram signals, 3-axial acceleration from the left-ankle sensor, 3-axial acceleration from the right-lower-arm sensor, 3-axial magnetometer from the left-ankle sensor, 3-axial magnetometer from the right-lower-arm sensor, 3-axial gyroscope from the left-ankle sensor, 3-axial gyroscope from the right-lower-arm sensor. The recorded activities are 12 and are performed by 10 volunteers. 
All sensors were recorded at a sampling rate of 50Hz.

Datasets were initially rescaled using per-channel normalization. Then, the signals have been divided in windows with an overlap between subsequent segments of 50\%.\\ For UCI HAR and MHEALTH, a window of $2.56s$ was used, while for Skoda, a window of $2s$.
Furthermore, in MHEALTH, the data relative to electrocardiogram signals have been removed as the other datasets had no data of this type.

After the pre-processing phase, the first 50\% of sensory measurements in each sample represents the \emph{input sequences}. Consequently, the \emph{temporally inverted sequence} of the input is used as the target sequence for the reconstruction task. The \emph{remaining sensory measurements} are considered as the target sequence for future prediction.

\subsection{Evaluation Metrics}

All clustering methods are evaluated by \emph{Clustering Accuracy (ACC)}~\cite{strehl2002cluster} and \emph{Normalized Mutual Information (NMI)}~\cite{vinh2010information}, widely used in unsupervised learning scenarios. Both metrics expect values in the range [0, 1], with 1 being the perfect clustering and 0 being the worst.

The Clustering Accuracy (ACC) takes a cluster assignment and a ground truth assignment and then finds the best matching between them.  We used the \emph{Hungarian algorithm}~\cite{kuhn1955hungarian} to compute the best mapping (see Equation~\ref{eq:ACC}).

\begin{equation}
    ACC = max_n \; \frac{ \sum_{i=1}^n 1\{ l_i=m(c_i) \} }{n}
\label{eq:ACC}
\end{equation}
where $l_i$ is the true label, $c_i$ the cluster assignment, and $m$ ranges over all possible one-to-one mappings between clusters and labels.

The Normalized Mutual Information (NMI), between ground-truth labels and the labels obtained by clustering, determines the class labels' entropy reduction, assuming the cluster labels are known. It is used for determining the quality of the clustering and is defined in Equation~\ref{eq:NMI}.
\begin{equation}
    NMI(y,c) = \frac{2 I(y,c)}{H(y)+H(c)}
    \label{eq:NMI}
\end{equation}
where $y$ is the true label, $c$ is the obtained cluster label, $I(.,.)$ is the mutual information and $H(.)$ is the entropy.

%% file: sections/results.tex
\section{Experiments Results}
\label{section:results}

\begin{table*}[h!]
\centering

\caption [Comparison of Clustering performances on HAR Datasets.]{Clustering performances in terms of NMI and ACC on the three datasets achieved by DISC and state-of-the-art methods. The first group of rows shows the performance of traditional clustering techniques applied on raw data. The second group of rows shows the results achieved by DISC and state-of-the-art AutoEncoders combined with traditional clustering techniques. The last group of rows shows the comparison between DISC and the end-to-end deep clustering methods. 
}

\resizebox{1\textwidth}{!}{
\begin{tabular}{r|cccc|cccc|cccc|cccc}
\toprule

& \multicolumn{4}{c}{\textbf{UCI HAR}}  & \multicolumn{4}{c}{\textbf{Skoda}} & \multicolumn{4}{c}{\textbf{MHEALTH}} & \multicolumn{4}{c}{\textbf{Average}} \\

& \multicolumn{2}{c}{\textbf{Train}} & \multicolumn{2}{c}{\textbf{Test}} & \multicolumn{2}{c}{\textbf{Train}} & \multicolumn{2}{c}{\textbf{Test}}& \multicolumn{2}{c}{\textbf{Train}} & \multicolumn{2}{c}{\textbf{Test}} & \multicolumn{2}{c}{\textbf{Train}} & \multicolumn{2}{c}{\textbf{Test}}\\

 &\textbf{NMI}&\textbf{ACC}&\textbf{NMI}&\textbf{ACC}&\textbf{NMI}&\textbf{ACC}&\textbf{NMI}&\textbf{ACC}&\textbf{NMI}&\textbf{ACC}&\textbf{NMI}&\textbf{ACC}&\textbf{NMI}&\textbf{ACC}&\textbf{NMI}&\textbf{ACC}\\
\midrule

\multicolumn{14}{l}{\textbf{Traditional Clustering on Input Data Space}} & & & \\[0.25cm]

$k$-Means     & 49.36 & 45.04 & 45.90 & 40.69 & 38.54 & 39.18 & 36.93 & 38.59 & 41.85 & 43.95 & 42.14 & 44.41  & 43.25 & 42.72 & 41.66 & 41.23 \\
AC-Average  & 1.18 & 19.16 & 1.8 & 18.29    & 1.51 & 13.88 & 18.38 & 14.37  & 12.12 & 10.35 & 9.63 & 10.19   & 4.93 & 14.46 & 9.94 & 14.28 \\
AC-Complete & 3.45 & 19.56 & 19.22 & 31.69  & 33.29 & 31.32  & 31.86 & 28.07 & 23.49 & 42.10 & 30.02 & 13.74 & 20.07 & 30.99 & 27.03 & 24.50 \\
AC-Ward     & 40.73 & 42.26 & 47.71 & 43.26 & 41.79 & 41.35 & 36.99 & 33.19 & 48.07 & 48.26 & 48.54 & 48.60  & 43.53 & 43.96 & 44.41 & 41.68 \\[0.2cm]


\midrule

\multicolumn{14}{l}{\textbf{Traditional Clustering on proposed Recurrent AutoEncoding Space }} & & & \\[0.25cm]

DSC $k$-Means     & 51.93 & 60.19 & 45.49 & 55.62 & 53.75 & 47.56 & 50.64 & 42.62 & 54.86 & 43.96 & 55.75 & 48.24 & 53.51 & 50.57 & 50.62 & 48,82 \\

DSC AC-Average  & 45.18 & 37.57 & 46.41 & 34.61 & 18.88 & 16.96 & 38.59 & 30.22 & 34.54 & 20.47 & 47.01 & 29.26 & 32.86 & 25.00 & 44.00 & 31.36  \\

DSC AC-Complete & 40.66 & 40.03 & 40.81 & 43.67 & 32.55 & 32.47 & 41.57 & 35.93 & 42.23 & 35.05 & 44.42 & 36.51 & 38.48 & 35.85 & 42.26 & 38.70 \\

DSC AC-Ward & 75.27 & 74.78 & 52.83 & 60.33 & 55.81 & 51.51 & 54.41 & 45.96 & 61.07 & 48.91 & 57.04 & 46.28 & 64.05 & 58.40 & 54.76 & 50.85 \\ 
\textbf{(ours)} DISC $k$-Means     & 54.84 & 55.27 & 48.27 & 54.62 & 50.65 & 47.83 & 47.87 & 43.10 & 65.14 & 55.27  & 64.88 & 54.04 & 56.88 & 52.76 & 53.67 & 50.59  \\
\textbf{(ours)} DISC AC-Average  & 43.82 & 39.17 & 41.26 & 35.12 & 29.74 & 28.82 & 29.63 & 24.12 & 30.80 & 13.69 & 31.24 & 13.16 & 34.79 & 27.23 & 34.04 & 24.13  \\
\textbf{(ours)} DISC AC-Complete & 37.77 & 44.12 & 38.91 & 40.62 & 35.48 & 34.62 & 44.13 & 37.31 & 46.58 & 36.22 & 50.21 & 30.24 & 39.94 & 38.32 & 44.41 & 36.06  \\ %
\textbf{(ours)} DISC AC-Ward     & 64.21 & 63.14 & 51.75 & 60.77 & 52.85 & 47.92 & 54.10 & 43.45 & 68.85 & 58.31 & 70.98 & 59.52 & 61.97 & 56.46 & 58.94 & 54.58 \\[0.2cm]

\midrule

\multicolumn{14}{l}{\textbf{End-to-End Deep Clustering}} & & & \\[0.25cm]

DEC~\cite{xie2016unsupervised}  & 55.57 & 49.93 & 55.20 & 49.10 & 46.86 & 41.02 & 46.58 & 41.08 & 50.52 & 43.71 & 51.68 & 45.11 & 50.98 & 44.89 & 51.15 & 45.10  \\
IDEC~\cite{guo2017improved}  & 55.76 & 51.78 & 54.43 & 51.02 & 49.65 & 46.70 & 48.93 & 40.56 & 51.28 & 42.59 & 52.34 & 44.76 & 52.23 & 47.02 & 51.90 & 45.45  \\
DCEC~\cite{guo2017deep}   & 52.77 & 48.39 & 53.12 & 48.88 & 42.90 & 39.56 & 45.51 & 42.24 & 44.15 & 23.51 & 45.13 & 24.29 & 46.61 & 40.15 & 47.92 & 38.47  \\

DSC ($k$-Means)~\cite{abedin2020towards}  & 64.75 & 64.54 & 61.58 & 61.28 & 56.91 & 50.97 & 57.01 & 50.28 & 62.65 & 57.19 & 63.06 & 56.85 & 61.43 & 57.56 & 60.55 & 56,13 \\

DSC (Ward)~\cite{abedin2020towards}   & 76.43 & 78.79 & 71.25 & 75.41 & \textbf{56.97} & 52.90 & \textbf{59.06} & 53.48 & 59.42 & 51.57 & 60.91 & 53.33 & 64.27 & 61.08 & 63.74 & 60.74 \\

\textbf{(ours)} DISC ($k$-Means)  & \textbf{85.72} & \textbf{92.08} & 80.81 & \textbf{87.99} & 52.93 & 54.45 & 52.93 & \textbf{54.45} & \textbf{68.15} & \textbf{53.96} & \textbf{69.38} & \textbf{54.83} & \textbf{68.93} & \textbf{66.83} & \textbf{67.71} & \textbf{65.76}  \\
\textbf{(ours)} DISC (Ward)     & 85.47 & 81.86 & \textbf{81.91} & 79.40 & 54.91 & \textbf{54.57} & 55.93 & 53.45 & 63.23 & 50.74 & 61.29 & 48.82 & 67.87 & 62.39 & 66.38 & 60.56  \\[0.2cm]


\bottomrule

\end{tabular}}

\label{tab:results_labels}
\end{table*}


Table \ref{tab:results_labels} shows the clustering performance in terms of NMI and ACC on the UCI HAR, Skoda, and MHEALTH datasets achieved by the proposed and state-of-the-art methods. 

We used as traditional clustering methods $k$-Means~\cite{hamerly2003learning} and Agglomerative Clustering (AC)~\cite{mullner2011modern}, with three different linkage types (i.e., Average, Complete, and Ward).

The first group of rows shows the performance of traditional clustering techniques applied on the raw data. 
The second group of rows shows the results achieved by our architecture DISC and state-of-the-art- AutoEncoders combined with traditional clustering techniques. The last group of rows shows the comparison of DISC method with the state of the art. 
Boldface font stands for best results. 

Overall the results confirm the effectiveness of deep clustering approaches against traditional clustering approaches on input space data and AutoEncoding space. On average, our proposal performs better with respect to the state-of-the-art.


DSC (Ward) performs slightly better (on average of about 3\%) than our proposal in terms of NMI only on the Skoda dataset. Our method overcomes the best method in the state of the art of about 10\% in terms of NMI and ACC on both UCI HAR  and MHEALTH. Improvements are mainly related to the use of convolutional GRU instead of simply GRU.

\begin{figure*}[h]
  \centering
  \includegraphics[width=0.8\textwidth]{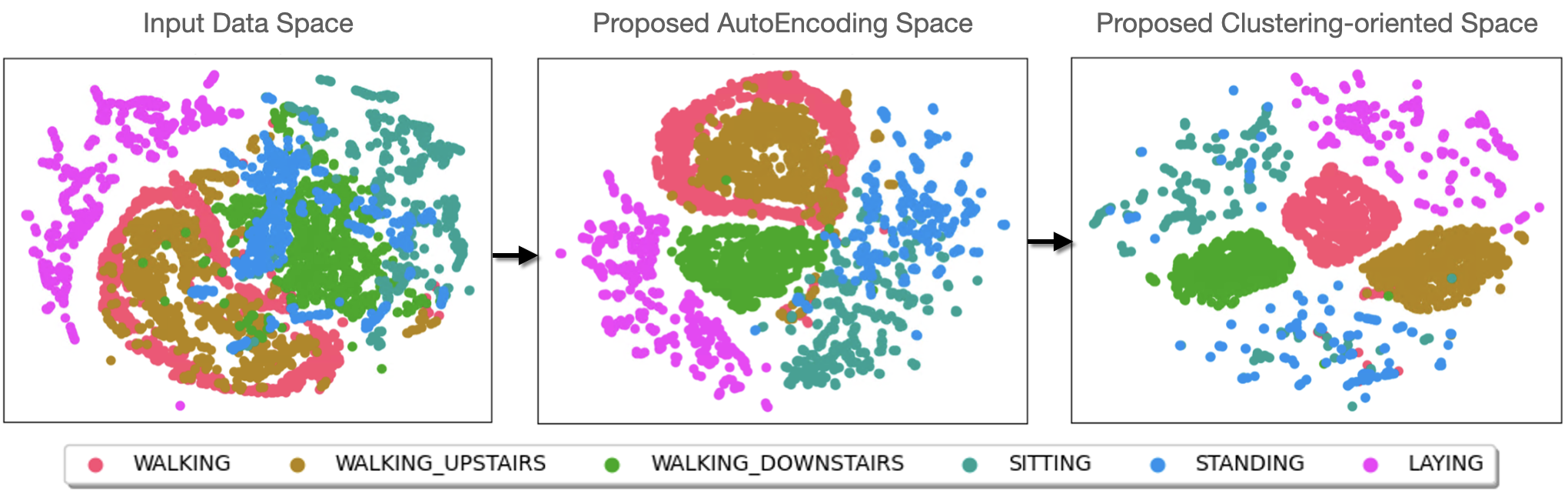}
  \caption{Feature space progression of the UCI HAR Dataset with DISC ($k$-Means initialization).}
  \label{fig:uci_DISC}
\end{figure*}

Figure~\ref{fig:uci_DISC} shows the progression of the learned feature space from the initial setup to the final clustering-oriented embedding space for the UCI HAR dataset. The progression is achieved by firstly using the Principal Component Analysis (PCA)~\cite{jolliffe2005principal} to reduce the embedding from 256 to 50 dimensions, and then by using the t-Distributed Stochastic Neighbor Embedding (t-SNE)~\cite{van2008visualizing} method applied for mapping 50-dimensional data into 2-dimensional data. 

The proposed architecture discovers well-defined and separated clusters of activity segments with strong correspondence to the ground-truth labels. In particular, three activities (walking, walking upstairs, and walking downstairs) are correctly delimited, while the others (sitting, standing, and lying) are well delimited but more dispersed, due to the high number of subjects in the dataset, which increase the inter-variability of activities.

%% file: sections/conclusions.tex
\section{Conclusions and Future Directions}
\label{section:conclusions}
The paper presents DISC, a deep-learning-based clustering architecture that learns highly discriminative spatio-temporal representations with reconstruction and future prediction tasks on multi-dimensional inertial signals. 
%
The architecture includes a recurrent encoder ConvGRU, two conditional decoders GRU, and a clustering criterion to predict unlabelled human activities-related signals. 


The proposed architecture has been compared with state-of-the-art approaches in both traditional and DL-based clustering approaches, demonstrating its effectiveness on three HAR datasets.

We plan to include DISC within the Continuous Learning Platform (CLP)~\cite{ferrari2019framework} framework. CLP is a platform that semi-automatically integrates heterogeneous labeled data and provides them in a homogeneous form. Integrating DISC into CLP would allow the dataset to be also populated with unlabeled signals that are collected with consumer devices.
Moreover, DISC can be also be employed in the early stages of a personalization strategy and thus allowing the recognition of ADL to a never seen user~\cite{amrani2021personalized}.

